\documentclass[conference]{IEEEtran}
\IEEEoverridecommandlockouts

\usepackage[square,numbers]{natbib}

\usepackage{amsmath,amssymb,amsfonts}
\usepackage{algorithmic}
\usepackage{graphicx}
\usepackage{textcomp}
\usepackage{xcolor}

\usepackage{hyperref}

\usepackage{times}
\usepackage{latexsym}
\usepackage{graphicx}
\usepackage{multicol}
\usepackage{soul}

\usepackage[T1]{fontenc}

\usepackage[utf8]{inputenc}

\usepackage{microtype}

\usepackage{inconsolata}

\def\BibTeX{{\rm B\kern-.05em{\sc i\kern-.025em b}\kern-.08em
    T\kern-.1667em\lower.7ex\hbox{E}\kern-.125emX}}
\begin{document}

\title{Conceptualizing Suicidal Behavior : Utilizing Explanations of Predicted Outcomes to Analyze Longitudinal Social Media Data}
\author{\IEEEauthorblockN{Van Minh Nguyen\IEEEauthorrefmark{1},
Nasheen Nur\IEEEauthorrefmark{2},
William Stern\IEEEauthorrefmark{2}, 
Thomas (Ty) Mercer\IEEEauthorrefmark{2}, Chiradeep Sen\IEEEauthorrefmark{3},\\
Siddhartha Bhattacharyya\IEEEauthorrefmark{2}, Victor Tumbiolo\IEEEauthorrefmark{2} and
Seng Jhing Goh\IEEEauthorrefmark{2}}
\IEEEauthorblockA{\IEEEauthorrefmark{1}Mathematical Sciences,
Florida Institute of Technology, Melbourne, FL 32901 USA}
\IEEEauthorblockA{\IEEEauthorrefmark{2}Electrical Engineering and Computer Science, Florida Institute of Technology, Melbourne, FL 32901 USA}
\IEEEauthorblockA{\IEEEauthorrefmark{3}Mechanical and Civil Engineering, Florida Institute of Technology, Melbourne, FL 32901 USA}
\thanks{Manuscript received July 31,2023; revised October 4, 2023. 
Corresponding author: Nasheen Nur (email: https://sites.google.com/view/nasheennur/home).}}





\maketitle

\begin{abstract}
The COVID-19 pandemic has escalated mental health crises worldwide, with social isolation and economic instability contributing to a rise in suicidal behavior. Suicide can result from social factors such as shame, abuse, abandonment, and mental health conditions like depression, Post-Traumatic Stress Disorder (PTSD), Attention-Deficit/Hyperactivity Disorder (ADHD), anxiety disorders, and bipolar disorders. As these conditions develop, signs of suicidal ideation may manifest in social media interactions. Analyzing social media data using artificial intelligence (AI) techniques can help identify patterns of suicidal behavior, providing invaluable insights for suicide prevention agencies, professionals, and broader community awareness initiatives. Machine learning algorithms for this purpose require large volumes of accurately labeled data. Previous research has not fully explored the potential of incorporating explanations in analyzing and labeling longitudinal social media data. In this study, we employed a model explanation method, Layer Integrated Gradients, on top of a fine-tuned state-of-the-art language model, to assign each token from Reddit users' posts an attribution score for predicting suicidal ideation. By extracting and analyzing attributions of tokens from the data, we propose a methodology for preliminary screening of social media posts for suicidal ideation without using large language models during inference.
\end{abstract}


\section{Introduction}

The alarming rise in mental health challenges, which the COVID-19 pandemic has exacerbated, has made early detection of suicidal ideation a pressing concern. Due to their widespread usage and user-generated content, social media platforms provide a valuable data source for detecting signs of mental distress. However, effectively utilizing this data to identify potential suicidal ideation poses numerous challenges.

While most current approaches focus primarily on detecting suicidal ideation within a given dataset, they often overlook the logistical aspects of collecting and analyzing the data \cite{pmid35805856,ARAQUE2017236}. For instance, with up to 50,000 monthly posts on suicidal/depression-related subreddit, with statistics collected for February 2023, and the platform's API limitations, it becomes challenging to scrape all submissions for analysis. Moreover, most existing techniques rely on large language models (LLMs) models, which can be computationally expensive and ill-suited for real-time analysis or deployment in resource-constrained environments.

Additionally, the black-box nature of many of these machine learning models makes it difficult for mental health professionals to understand and trust the predictions. This lack of transparency can hinder the adoption and use of these models in real-world settings, limiting their practical impact on mental health diagnosis and intervention.

We propose a cost-effective method to identify potential suicidal ideation from social media data to address the challenges mentioned above. Using token attribution (the details of which are discussed in the methodology section) extracted from the large language models (LLMs), we can classify users' long-context social media posting history without relying on larger context models. The approach reduces the computational requirements and time needed to analyze social media data, making it more feasible for large-scale and real-time applications. Furthermore, by focusing on explainablenable results backed by large language models, our method can help mental health professionals better understand and trust the predictions, ultimately facilitating more effective diagnosis and intervention strategies.

This paper has three major contributions: 
\begin{itemize}
    \item Utilized Layer Integrated Gradients and token attributions from LLMs as interpretable features for predicting suicidal ideation in social media posts without using LLMs in the loop for inference.
    \item Introduced TF-IDF scaling with token attributions and evaluated its effectiveness in predicting suicidal ideation
    \item Evaluated the efficacy of this method for predicting suicidal ideation in longitudinal data. In the context of data analysis, ``longitudinal data'' refers to data aggregated over time from the same users.
\end{itemize}


\section{Related Work}



In this section, we discuss the background of Large Language models and social media data analysis from generic applications of NLP to specific domains, primarily in the mental health domain. 
\subsection{Large Language Models for Deep Network Analysis in NLP}

The advent of Encoder-Based Models, especially the Bidirectional Encoder Representations from Transformers (BERT), marked a turning point in Natural Language Processing (NLP) \cite{devlin2018bert, alaparthi2020bidirectional}. These models capture complex linguistic contexts and dependencies, significantly advancing various NLP tasks. BERT, unlike its predecessors, which facilitated unidirectional or shallow bidirectional contextual representations, excels due to its deep bidirectional understanding \cite{devlin2018bert}. This denotes that the model concurrently processes the entire sequence of words, which leads to a more encompassing comprehension of the text.

To utilize BERT for sequence classification tasks, a unique strategy involving a special token, [CLS], is employed. This token is appended at the beginning of each input sequence during pre-training, but it does not carry any classification-specific information at this stage. It is during the fine-tuning stage for specific tasks like text classification that the final hidden state corresponding to the [CLS] token becomes vital. This hidden state serves as the aggregate sequence representation and is input to a fully connected layer for predicting the sequence's class label. In essence, BERT can classify the text based on the information encoded in this [CLS] token.

A crucial aspect of these models' success is the ability to adapt pre-trained models to specific domains and tasks, as demonstrated in \cite{gururangan2020don}. This work showcased the utility of domain-specific pretraining, resulting in performance gains in both high- and low-resource settings. Adapting to the task's unlabeled data improves performance even after domain-adaptive pretraining. The authors show that adapting to a task corpus augmented using simple data selection strategies is an effective alternative, especially when resources for domain-adaptive pretraining might be unavailable. In the mental health domain, domain-specific language models are pretrained and released, which facilitates the early detection of mental health conditions. \citeauthor{ji2023domain} \cite{ji2023domain} described domain-specific continued pretraining of language models for capturing long context which facilitates the early detection of mental health conditions.

Building on this foundation, specialized pre-trained models such as MentalBERT and MentalRoBERTa emerged, tailored for mental health applications \cite{ji2021mentalbert}. According to the authors, these models are pre-trained on mental health-related subreddits, improving their performance on mental disorder detection tasks. Additionally, other research  introduced MentalLongformer \cite{ji2023domain} to address the needs of long-context text prevalent in social media posts, extending the capabilities of traditional LLMs to include long-sequence modeling for mental health.

Attributing the prediction of deep networks to their input features is another critical research area \cite{sundararajan2017axiomatic,zhang2021fine,hao2021self,senior2020improved}. \citeauthor{sundararajan2017axiomatic}\cite{sundararajan2017axiomatic} introduced Integrated Gradients, an axiomatic approach that assigns importance to input features. The authors identify two fundamental axioms---Sensitivity and Implementation Invariance that attribution methods ought to satisfy. Further extending this idea, \citeauthor{hao2021self} \cite{hao2021self} implemented Layer Integrated Gradients for Transformers, allowing for token-level attribution and understanding of information interactions within Transformers. The authors extract the most salient dependencies in each layer to construct an attribution tree, which reveals the hierarchical interactions inside Transformer. This implementation is made accessible through the Transformer-Interpret library \footnote{\url{https://github.com/cdpierse/transformers-interpret}}.

Understanding the effect of context on word similarity is another pivotal aspect in this field. Bidirectional Encoder Representations from Transformers (BERT) can generate a deeper understanding of the text than unidirectional models, such as: term frequency-inverse document frequency (TF-IDF) \cite{devlin2018bert}. \citeauthor{chen2020ferryman} \cite{chen2020ferryman} proposed a unique approach of integrating token-level TF-IDF weighting with BERT, showing improved performance in English. TF-IDF for attribution is a precursor to its use for classification. The authors showed that with the TF-IDF score, it is possible to calculate the relevance between a word and a specific document without taking into account the position of a word in a sentence.

\subsection{Social Media Data Analysis for Mental Health Care}

Social media data analysis has gained significant attention in mental health care. By analyzing social media data, researchers and clinicians can gain insights into individuals' mental health states, identify patterns, and develop strategies for intervention and support \cite{de2013predicting, coppersmith2015adhd,guntuku2017detecting}. Researchers offer a starting point for understanding the application of social media data analysis in mental health care and showcase different methodologies, techniques, and insights from analyzing social media data to improve mental health support and intervention \cite{saxena2022explainable, zhang2023depression,lamichhane2023evaluation, sarkar2022predicting, nagahisarchoghaei2023empirical}. \citeauthor{de2013predicting} \cite{de2013predicting} analyze the language and behavioral changes in the social media posts of women who recently gave birth to identify signs of postpartum depression. Another exciting research by \citeauthor{ernala2017linguistic} \cite{ernala2017linguistic} examines the relationship between language patterns in social media disclosures of mental illness and therapeutic outcomes. The authors use machine learning techniques to analyze linguistic markers and predict outcomes, aiming to provide insights into the potential of social media for mental health support. Multiple research works provide further insights into the challenges faced in utilizing social media data for mental health analysis and highlight the need for careful consideration of ethical concerns, data quality, and generalizability limitations \cite{chancellor2020methods,calloway1993accuracy,de2017language}.

In this paper, we study the utilization of token attribution in predicting suicidal ideation without LLM during inference, with an objective to reduce the overhead in preliminary screening of the suicide risk from social media posts. We also expand upon this idea by implementing TF-IDF scaling on token level, and evaluating the effects on prediction performance. Finally, we implement this novel method to predict longitudinal and long context-length (usually more than 4096 tokens, the longest context length for a non-autoregressive model - Longformer \cite{beltagy2020longformer}), and evaluate the method's performance in predicting suicidal ideation in this case.

\section{Methodology}

\begin{figure}[htbp]
    \centering
    \includegraphics[width=0.4\textwidth]{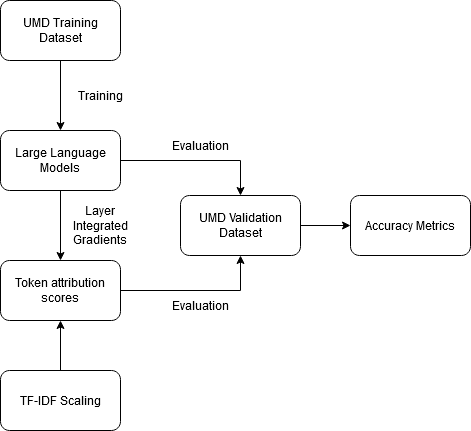}
    \caption{Workflow diagram of methods used}
    \label{fig:workflow-diag}
\end{figure}

The code for data processing, fine-tuning on cluster, and analysis of token attributions are publicly available on GitHub \footnote{\url{https://github.com/FIT-suicide-prevention-research/token-attribution-analysis}}. Following subsections explains the workflow of our approach (See Figure \ref{fig:workflow-diag}).

\subsection{Datasource and Processing}

The datasets employed in this study include the UMD dataset \cite{zirikly2019clpsych,zirikly2019clpsych} for model fine-tuning and evaluating our methodologies' performance. The University of Maryland Reddit Suicidality Dataset Version 2 (UMD dataset) is a private dataset curated to aid research on suicidality and suicide prevention. It is derived from the 2015 Full Reddit Submission Corpus, specifically from the r/SuicideWatch subreddit, with users anonymized to maintain privacy. The dataset has been annotated by experts and crowdsourcers on a four-point scale to classify the level of suicide risk (no risk, low, moderate, and severe risk). The dataset includes these sub-datasets:
\begin{itemize}
        \item crowd: Annotations by volunteers (crowdsourced annotations) for 621 users who posted on the "SuicideWatch" subreddit and 621 control users, splitted into a train and validation test to validate during training of the model.
        \item expert: Annotations by experts for 245 users who posted on SuicideWatch and 245 control users. We'll be using this "gold dataset" to evaluate (test) the efficacy of the model
\end{itemize}

Risk levels are designated as a, b, c, d, or None. Risk classes a, b, c, d corresponded to no, low, moderate, and severe risk; None is the default label for control users. 

Each sub-datasets are provided as 2 CSV (comma-separated values) files: one containing users' anonymized IDs, their posts on Reddit and the corresponding subreddits (including the users' posts on non-mental-illness related subreddits); one containing users' IDs and suicidal risk levels (either by experts or crowdsourced). Both these data files are first inner joined by userID before any other preprocessing steps are taken.

For generating the fine-tuning dataset, we filtered user posts only from /r/SuicideWatch, and classify each posts individually without the effect of user ID. The risk level are remapped from a, b, c, d into numerical values 0, 1, 2, 3 for compatibility with the API of the HuggingFace\footnote{\url{https://github.com/huggingface/transformers}}) Transformer library.

The longitudinal dataset is generated by concatenating posts from each users on all subreddit they posted, ordered by timestamp. For all generated datasets, we remove all rows with text length less or equal to 1 to simplify the downstream fine-tuning/inference tasks


\textbf{Ethical consideration.} To ensure the highest standards of ethical conduct in our research, all authors have undergone comprehensive training. Specifically, every author has completed the "Social \& Behavioral Research" as well as the "Social and Behavioral Responsible Conduct of Research" certifications from the Citi Program. Furthermore, all Principal Investigators have been trained in the Health Insurance Portability and Accountability Act (HIPAA) protocols. The first author, in addition to the aforementioned qualifications, has secured the "Data or Specimens Only Research" certification from the Citi Program, which encompasses HIPAA training as an integral component.

Our research received approval from the Institutional Review Board (IRB). Moreover, dataset access was reviewed and granted by Dr. Philip Resnik from University of Maryland, the proprietor of the dataset in use. These approvals and certifications underscore our commitment to maintaining the confidentiality, privacy, and ethical handling of the data used in our study.

\subsection{Model fine-tuning and token attribution}

We first conducted an exploratory comparison of various encoder-based LLMsfor the classification task using the UMD dataset. The models compared included MentalBERT (large, base, cased, and uncased) and MentalRoBERTa (large and base), which are BERT/RoBERTa models pre-trained on mental health-related subreddit submissions \cite{ji2021mentalbert}. We also included BERT (cased and uncased, base and large) \cite{devlin2018bert} as an original model comparison with MentalBERT models. Other models like Microsoft DeBERTa v3 (large and base) \cite{he2023debertav3} and XLM-RoBERTa-base \cite{conneau2020unsupervised} are also assessed as they are BERT derivative models with minor architecture revisions and training paradigms. 
However, models specifically designed for longer context classification such as MentalLongformer and MentalXLNet \cite{ji2023domain} ran out of memory on our NVIDIA A100 40GB during fine-tuning, and thus are not included in this stage of analysis.

In terms of data pre-processing for parsing into language models, we chose to leave the reddit post texts in their original state without any cleaning, as the cleaning process is done implicitly within the model's tokenizer. The tokenization method for the BERT and BERT based models is WordPiece \cite{devlin2018bert}, a tokenization method where words are divided into a limited set of common sub-words ("wordpieces") based on frequency of occurences and decide its merge strategy by maximizing the likelihood of the sub-words pair\cite{mike2012googles,wu2016googles}. For RoBERTa model, the tokenization method is byte-level Byte-Pair Encoding (BPE) \cite{liu2019roberta}. While BPE is more primitive than WordPiece where it only uses highest frequency of the subwords pair occurences \cite{sennrich2016neural}. Each LLM is fine-tuned on the UMD training set and subsequently evaluated on the UMD validation/test set. The BERT and RoBBERTa models we utilized are only capable of natively processing texts that are 512 tokens in length or less. In order to take the entire post history of each user into account, many of which are larger than 512 tokens, we implemented a sliding window method, introduced in \citeauthor{wang2019multipassage} \cite{wang2019multipassage}, to allow the models to read text larger than the 512 token  limit. For the optimization process, we used AdamW optimizer (the default optimizer for LLM fine-tuning provided by HuggingFace transformer library  with a learning rate of $10^{-5}$. We used a batch size of 8 for large models and 16 for base models. The training of each model spanned 5 epochs (as documented in \citeauthor{devlin2018bert} \cite{devlin2018bert}), saving the models as checkpoints every 10 steps. Fine-tuning is performed using an NVIDIA A100-40GB graphics processing unit. The model with the best validation accuracy is selected from these checkpoints.

Using the labeled text as predicted from the language model, we are able to generate model explanations using Layer Integrated Gradients by computing each token's attribution towards the predicted label. This technique results in attribution scores for each token in every position for multiple labels. Due to the characteristic functioning of transformers, each token (even in different positions) has a different attribution score. Finally, we aggregated all attribution scores for specific token across the entire corpus, yielding a list of all attributions for each label. To garner a better understanding of the relationships between tokens and their attributions to the predicted labels, we explored histogram plots for some tokens with high recurrences (shown in Figure \ref{fig:attribution-histogram}).

\section{Results}

\subsection{Comparison of different Large Language Models for social media suicidal ideation detection}

\begin{table}[tbp]
        \centering
        \begin{tabular}{lcc}
                \hline
                \textbf{Model}           & \textbf{Accuracy} & \textbf{Loss} \\
                \hline
                MentalRoBERTa-large      & \textbf{63.05\%}  & \textbf{1.01} \\
                MentalRoBERTa-base       & \textbf{63.05\%}  & 1.04          \\
                MentalBERT-large uncased & 62.56\%           & 1.04          \\
                MentalBERT-base cased    & 61.57\%           & 1.02          \\
                MentalBERT-base uncased  & 61.57\%           & 1.04          \\
                DeBERTa v3 Large         & 62.06\%           & 1.01          \\
                DeBERTa v3 Base          & 58.12\%           & 1.07          \\
                BERT-large uncased       & 60.59\%           & 1.08          \\
                BERT-large cased         & 59.60\%           & 1.05          \\
                BERT-base uncased        & 59.11\%           & 1.09          \\
                BERT-base cased          & 58.62\%           & 1.07          \\
                XLM-RoBERTa base         & 53.69\%           & 1.12          \\
                \hline
        \end{tabular}
        \centering
        
        \caption{Validation results after fine-tuning of various models on UMD training set.}
        \label{tab:fine-tuning}
\end{table}

The result of the fine-tuning models is shown in Table \ref{tab:fine-tuning}. MentalRoBERTa-large and MentalRoBERTa-base achieved the highest validation accuracy of 63.05\% and the lowest validation loss of 1.01 (the loss metric used for the sequence classification validation is multi-class cross-entropy loss).  From the result table we observe an interesting trend for BERT-type model: uncased model often outperforms cased model, except for model pretrained on social media data (MentalBERT). We believe this is due to the fact that the UMD dataset is collected from social media, where people often use informal language and do not capitalize their words. As such, the cased model cannot capture the nuances of the text, resulting in lower performance for the original models. However, this is not the case for MentalBERT cased models, which is pretrained on social media data and adapted for less capitalized, informal text.

We found that the large model and base model achieve a similar performance, with the exception of DeBERTa v3. This is likely due to the fact that the vocabulary size of the social media post is much smaller than that of the general text, resulting in less need for the larger vocabulary of bigger models. DeBERTa v3 exception is not the norm as it is pretrained on less data than other models but is still able to achieve similar performance to the top models, which shows that DeBERTa v3 improvement in disentangled attention and enhanced mask decoder is effective in improving the model performance. This prompts a future research direction of pretraining DeBERTa v3 on social media data (from checkpoints) to see if it can achieve better performance than other models.

As both MentalRoBERTa-large and MentalRoBERTa-base achieved the highest validation accuracy, and MentalRoBERTa-base has a smaller model size and lower inference time, we choose MentalRoBERTa-base as our model for further analysis.

\subsection{Transformer Explanations and Visual Analysis}

\begin{figure*}[!h]
    \centering
    \includegraphics[width=0.8\textwidth]{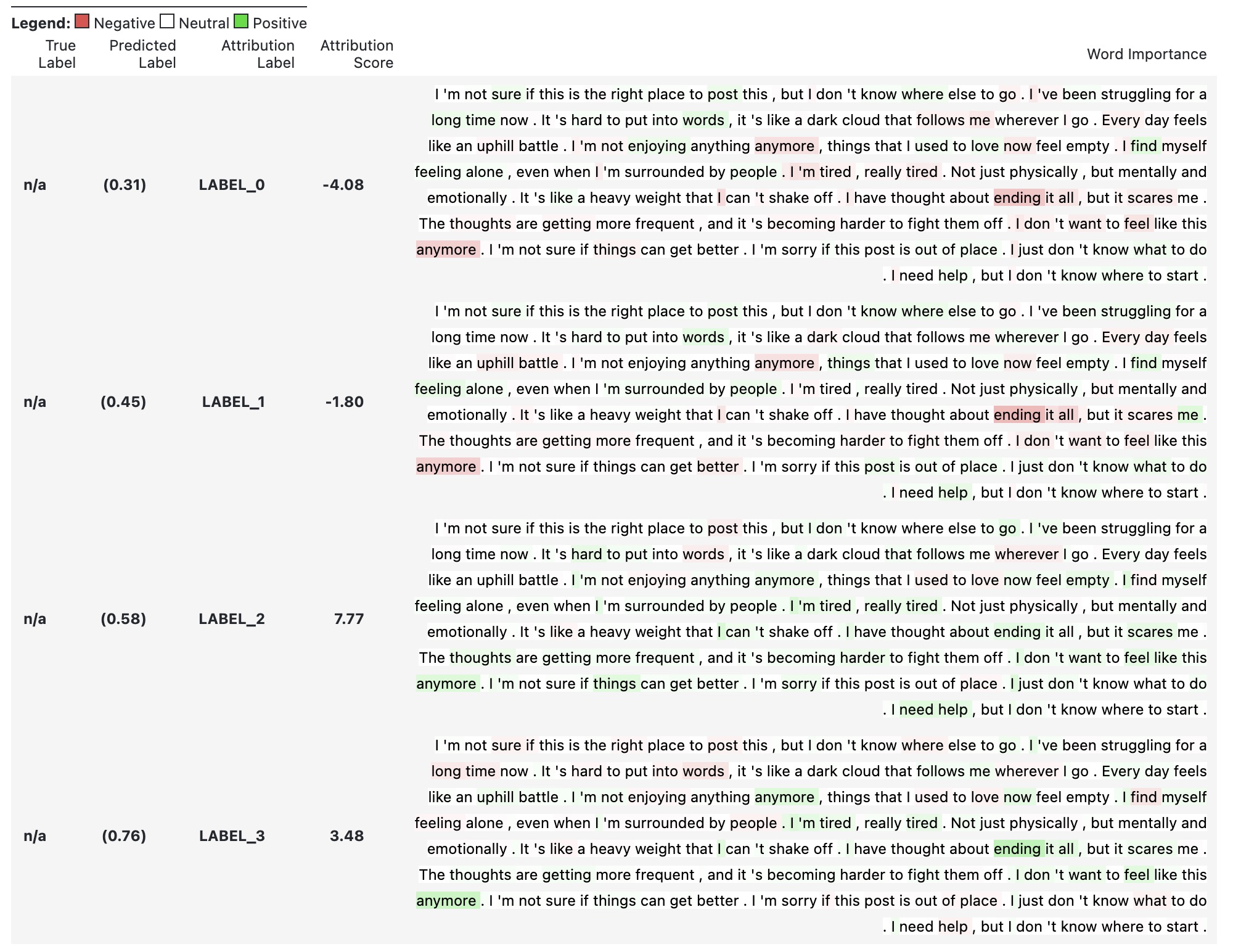}
    \caption{Example of transformer-interpret visualization. The example text is generated by ChatGPT4 and is confirmed not in UMD dataset or Reddit /r/SuicideWatch submission history. This example is classified as label C (LABEL\_2) by the model, which is medium risk of suicide according to UMD dataset}
    \label{fig:transformer-interpret}
\end{figure*}

\begin{figure*}[!h]
    \centering
    \begin{multicols}{2}
    \includegraphics[width=0.8\linewidth]{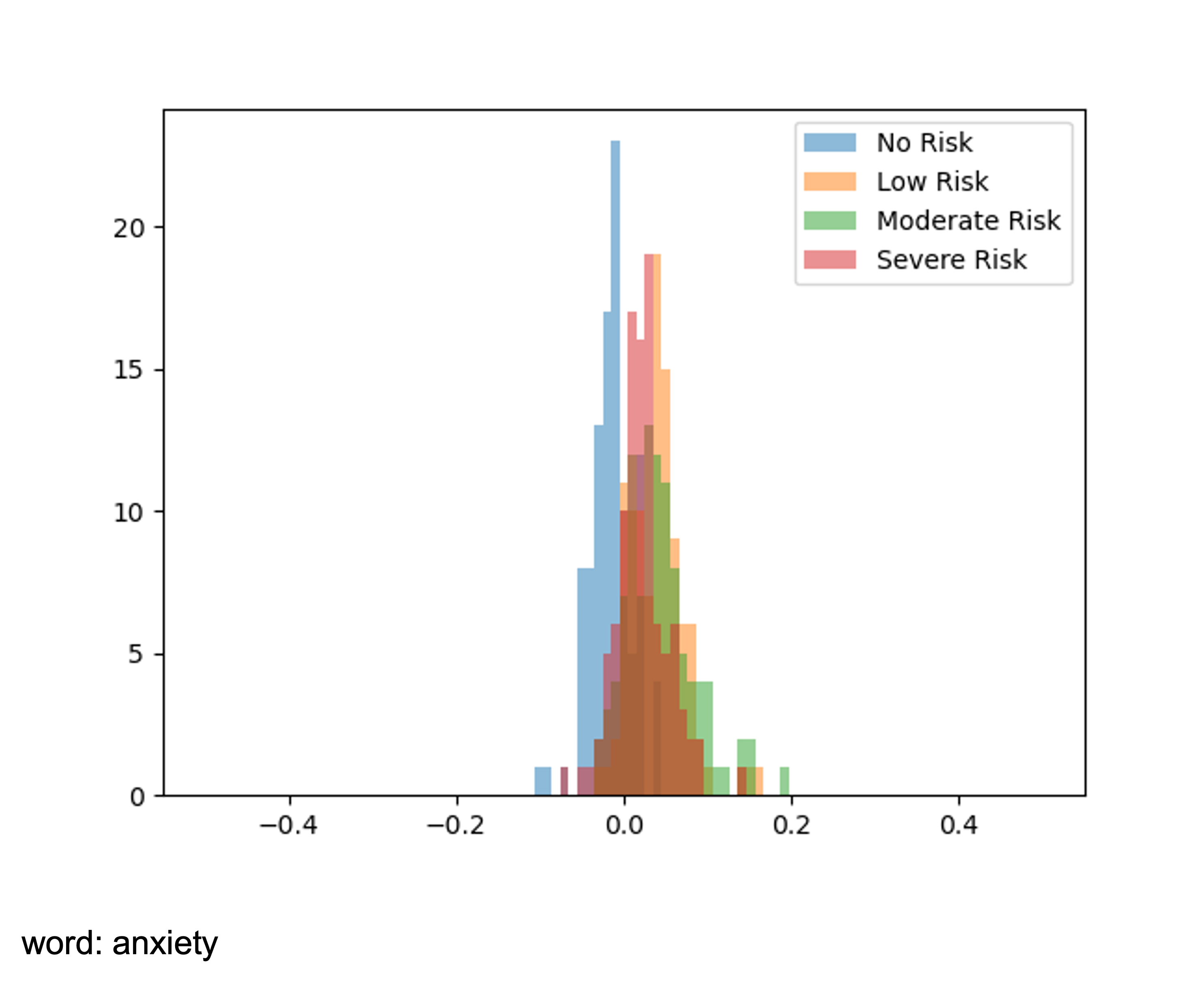}\par 
    \includegraphics[width=0.8\linewidth]{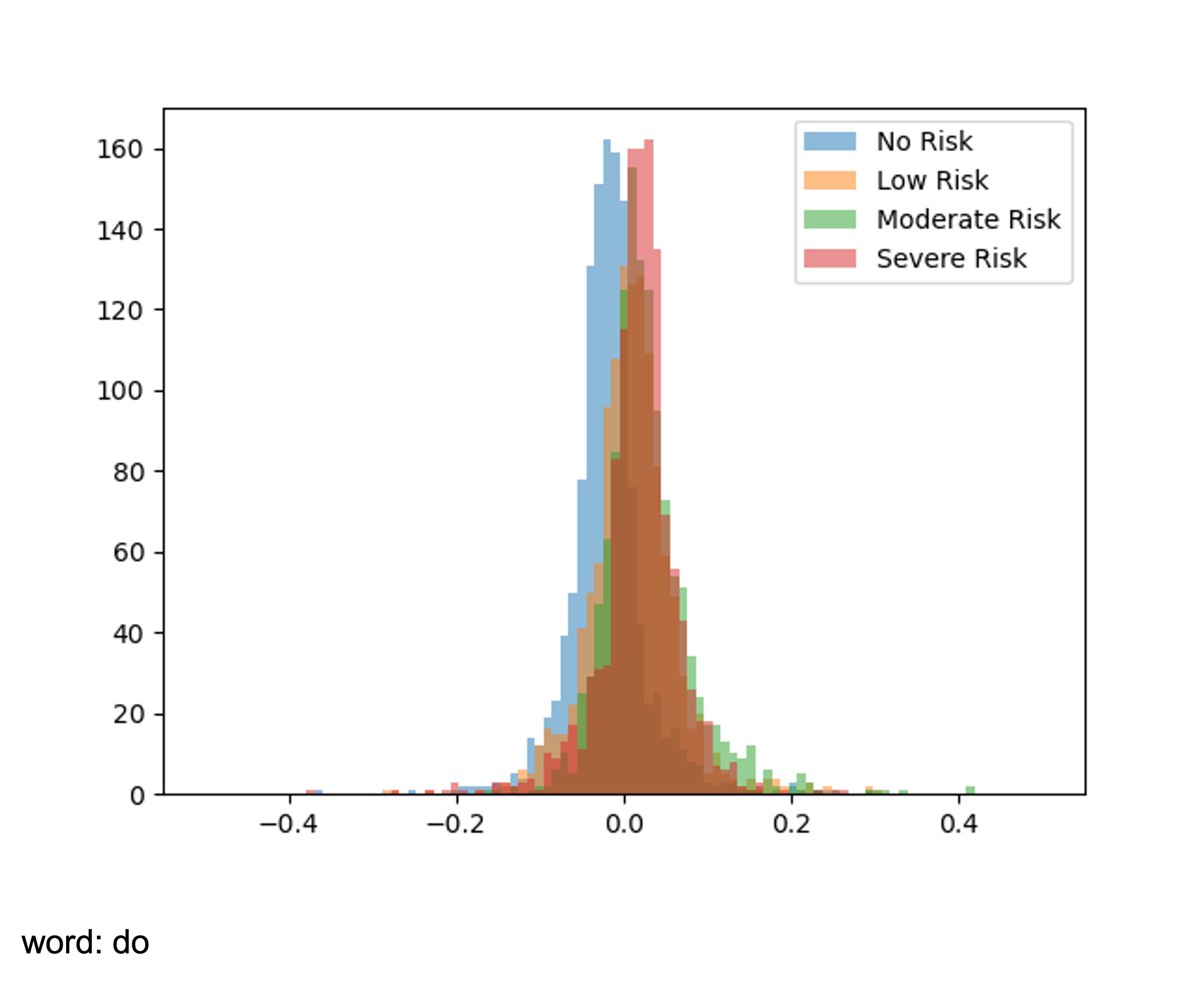}\par 
    \end{multicols}
    \begin{multicols}{2}
            \includegraphics[width=0.8\linewidth]{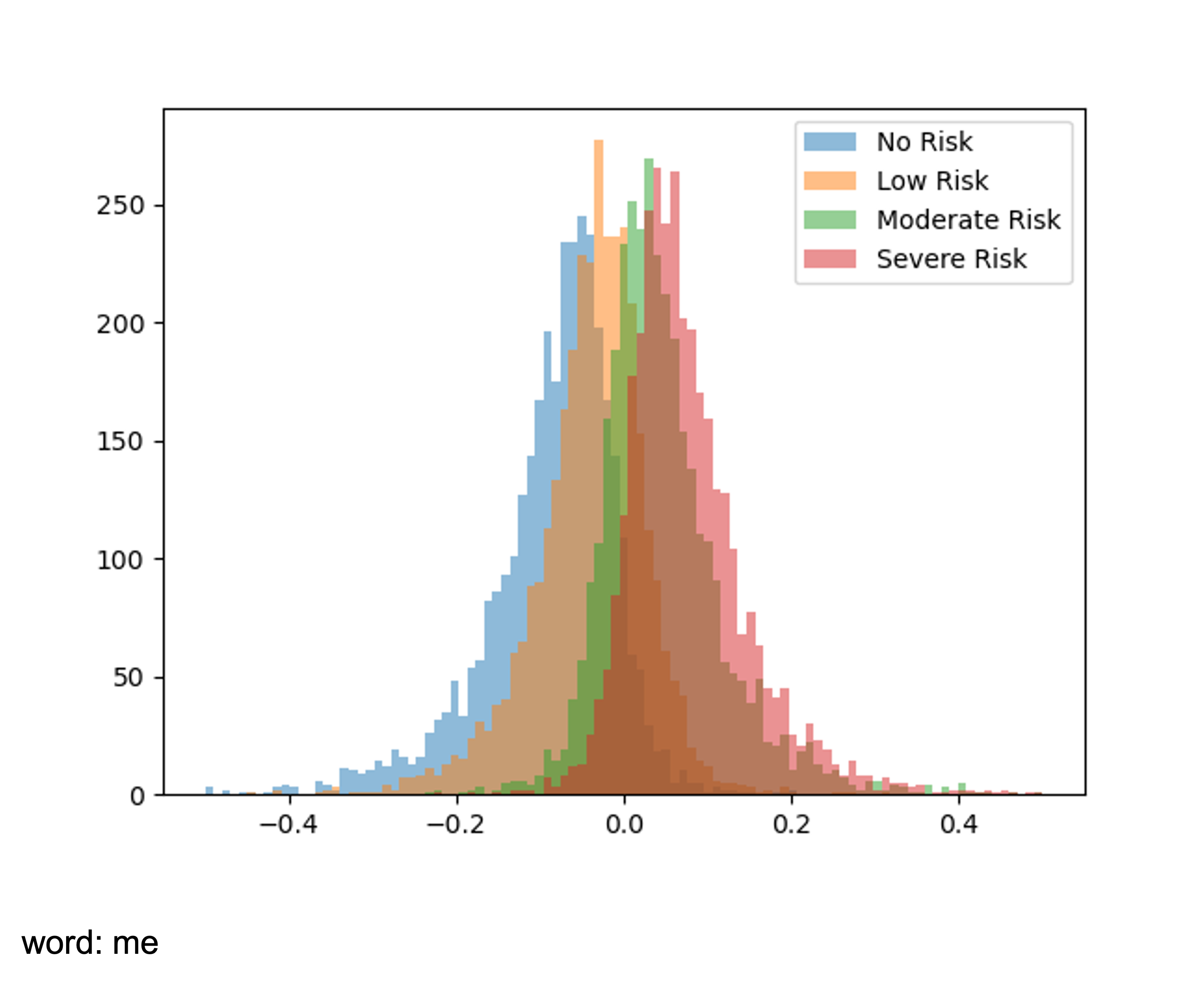}\par 
    \includegraphics[width=0.8\linewidth]{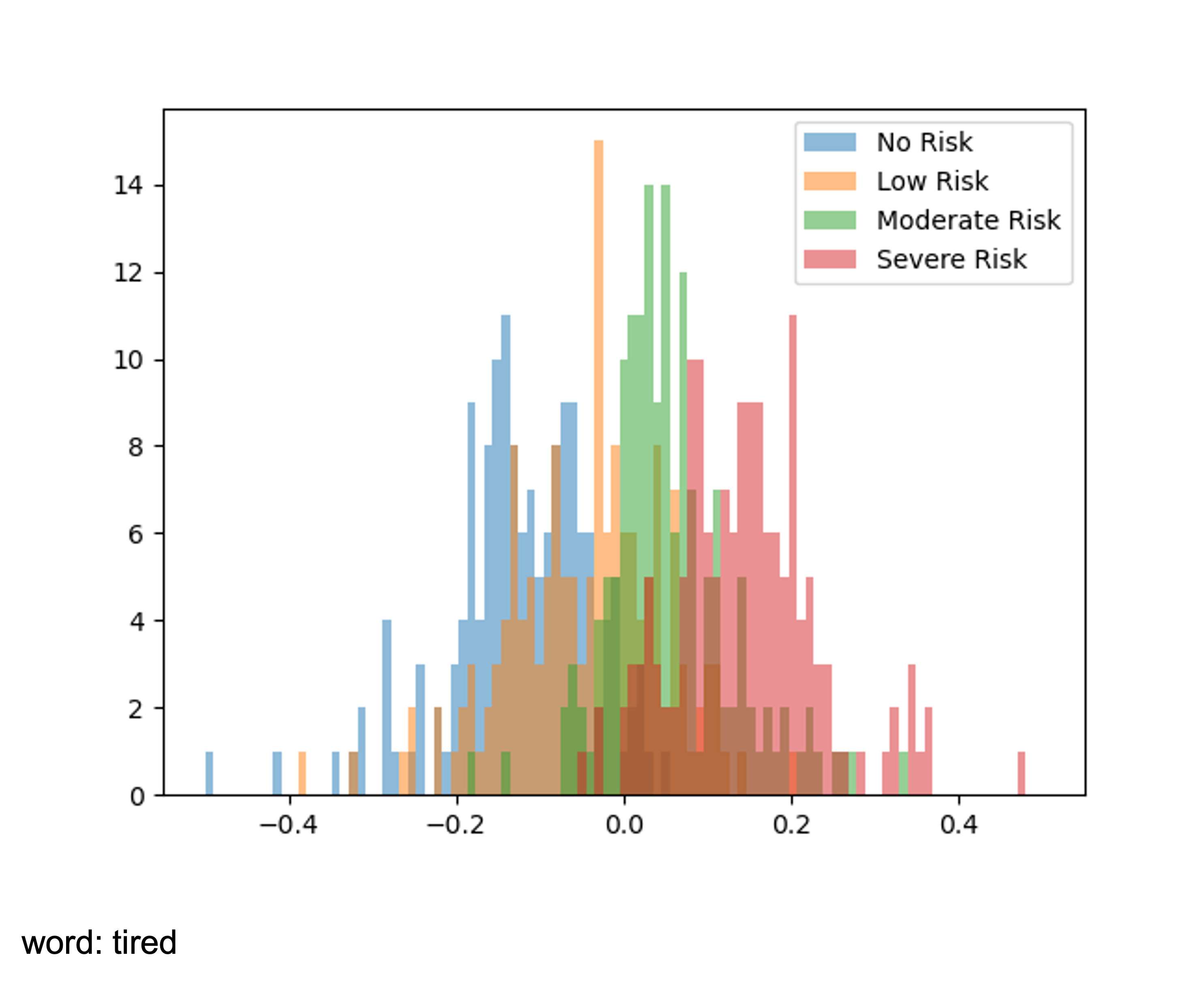}\par 
    \end{multicols}
    \caption{Sample token attribution histogram}
    \label{fig:attribution-histogram}
\end{figure*}

Layer Integrated Gradients output attribution for each token in each position of the document (social media post). Using the transformer-interpret library we are able to output visualizations of each token's attribution and their contribution towards the final prediction (Figure \ref{fig:transformer-interpret}). While each attribution score is unique to each token, we can combine all attribution scores for a specific token over the entire corpus, creating a histogram of attributions for all tokens for each label (as shown in Figure \ref{fig:attribution-histogram}).

As shown in Figure \ref{fig:attribution-histogram}, tokens are distributed over ground truth labels (e.g., 0,1,2,3), and their statistics are used to determine the "representative value" of each attribute.

\subsection{Using token attribution for suicidal ideation detection}


\subsubsection{Suicidal ideation detection from /r/SuicideWatch}

\begin{table}[!tbp]
\centering
\begin{tabular}{lcc}
\hline
\textbf{Method} & \textbf{Recall} & \textbf{F1-score} \\ 
\hline
MentalRoBERTa-base* & 57.96 & 58.58 \\
Token attribution & {} & {} \\
\quad median attribution & 25.07 & 10.51 \\
\quad \quad + TF-IDF scaling & 26.20 & 16.05 \\
\quad mean attribution & 25.07 & 10.52 \\
\quad \quad + TF-IDF scaling & \textbf{26.76} & \textbf{16.63} \\
\hline
\end{tabular}
\caption{No-Low-Medium-High risk classification result. The baseline data is from original MentalBERT paper\cite{ji2021mentalbert}, and our results is from classifying expert labeled, gold dataset from UMD dataset}
\label{tab:all-risk}
\end{table}

We can see there's a huge gap between our method and the baseline result. However, the results from MentalBERT paper used a different method of relabeling for better results: they included control users, and relabeled every records to have 3 label classes without mentioning any details of how they relabeled them. We only put the results here as a reference and will not use it for our analysis going onwards. From the result table \ref{tab:all-risk}, we observe that recall score is much higher than F1-score. Depending on the research question, recall statistics are more favorable, for example, we want to correctly identified suicidal post and don't want mis-classify suicidal post as non-suicidal. 
One of our research questions is what effect TF-IDF scaling has on the token attribution. We hypothesized that adding TF-IDF scaling improves attribution performance by balancing frequently used terms. After implementing the TF-IDF scaling, our hypothesis seems to be supported as shown by higher F1-score when using TF-IDF scaling. Considering all options and ablations, mean attribution score scaled with TF-IDF provides the most balanced result between Recall and F1-score (as illustrated by previous result and Tables \ref{tab:no-any-risk} and \ref{tab:non-inter-risk}). This is a great news as considering computational complexity of calculating mean (always $O(n)$) and median (on average $O(n)$, worst case $O(n^2)$), using mean attribution lowers the classification time.

\begin{table}[!h]
\centering
\begin{tabular}{lcc}
\hline
\textbf{Method} & \textbf{Recall} & \textbf{F1-score} \\ 
\hline
Token attribution & {} & {} \\
\quad median attribution & \textbf{89.58} & 84.65 \\
\quad \quad + TF-IDF scaling & 88.73 & 85.90 \\
\quad mean attribution & \textbf{89.58} & 85.17 \\
\quad \quad + TF-IDF scaling & 89.01 & \textbf{87.01} \\
\hline
\end{tabular}
\caption{No risk - Any risk}
\label{tab:no-any-risk}
\end{table}

\begin{table}[!h]
\centering
\begin{tabular}{lcc}
\hline
\textbf{Method} & \textbf{Recall} & \textbf{F1-score} \\
\hline
Token attribution & {} & {} \\
\quad median attribution & 70.42 & 59.53 \\
\quad \quad + TF-IDF scaling & 70.14 & 63.99 \\
\quad mean attribution & 70.42 & 59.53 \\
\quad \quad + TF-IDF scaling & \textbf{70.99} & \textbf{65.14} \\
\hline
\end{tabular}
\caption{No-Low vs Medium-High risk}
\label{tab:non-inter-risk}
\end{table}

\subsubsection{Longitudinal Suicidal ideation detection from users' Reddit submission history}

As most papers we found using SOTA LLMs to predict outcome of UMD dataset only uses posts/corpus from /r/SuicideWatch, or predicting posts individually (due to models' context length limit) \cite{ji2023domain,ji2021mentalbert,chua-etal-2022-unified,lee2021micromodels}, we don't have any baseline to compare here. The longitudinal detection is processed by concatenating posts by each user, then classify the long text using our methodology.

\begin{table}[!tbp]
\centering
\begin{tabular}{lcc}
\hline
\textbf{Method} & \textbf{Recall} & \textbf{F1-score} \\ 
\hline
Token attribution & {} & {} \\
\quad median attribution & \textbf{18.37} & {6.29} \\
\quad \quad + TF-IDF scaling & {17.55} & {6.65} \\
\quad mean attribution & {17.55} & {6.05} \\
\quad \quad + TF-IDF scaling & {17.55} & \textbf{6.66} \\
\hline
\end{tabular}
\caption{No-Low-Medium-High risk longitudinal classification result}
\label{tab:all-risk-long}
\end{table}

Similar to previous results, table \ref{tab:all-risk-long} shows the similar result where TF-IDF scaling improves classification results. However, Recall score is hugely affected by this method. Regardless, the result is not up to par for classifying all 4 classes of risks. On the other hand, grouping classes shows great performance in both recall and F1-score (Tables \ref{tab:no-any-risk-long} and \ref{tab:non-inter-risk-long})

\begin{table}[!h]
\centering
\begin{tabular}{lcc}
\hline
\textbf{Method} & \textbf{Recall} & \textbf{F1-score} \\ 
\hline
Token attribution & {} & {} \\
\quad median attribution & \textbf{85.31} & \textbf{78.54} \\
\quad \quad + TF-IDF scaling & \textbf{85.31} & \textbf{78.54} \\
\quad mean attribution & \textbf{85.31} & \textbf{78.54} \\
\quad \quad + TF-IDF scaling & \textbf{85.31} & \textbf{78.54} \\
\hline
\end{tabular}
\caption{No risk - Any risk longitudinal}
\label{tab:no-any-risk-long}
\end{table}

\begin{table}[!h]
\centering
\begin{tabular}{lcc}
\hline
\textbf{Method} & \textbf{Recall} & \textbf{F1-score} \\
\hline
Token attribution & {} & {} \\
\quad median attribution & 64.90 & 51.81 \\
\quad \quad + TF-IDF scaling & 61.22 & 51.69 \\
\quad mean attribution & 64.08 & 51.40 \\
\quad \quad + TF-IDF scaling & \textbf{61.63} & \textbf{52.99} \\
\hline
\end{tabular}
\caption{No-Low vs Medium-High risk longitudinal}
\label{tab:non-inter-risk-long}
\end{table}

\section{Conclusion}

In this work, we conducted an exploratory comparison of various encoder-based large language models (LLMs) for detecting suicidal ideation from social media posts, with a focus on the Reddit platform. Our approach employed the UMD dataset training data for model fine-tuning and analyzed the efficacy of the models on the "gold dataset" manually labelled by experts in the same UMD dataset. We also incorporated a novel method for transformer explanations and visual analysis using Layer Integrated Gradients, thus achieving interpretability for our model's predictions.

The MentalRoBERTa-base model demonstrated the best performance in terms of accuracy and is therefore selected for further analysis. We then integrated token attribution statistics in our approach to enhance model interpretability. Preliminary results for suicidal ideation detection indicate a substantial gap in performance when comparing to baseline results, emphasizing the potential for improvement. However, if we group the labeled data into groups, the result of those merged class is significantly better than the original label. This could be due to a bad segregation/decision boundary between classes of similar risk-level, for example, taking a look at Figure \ref{fig:attribution-histogram}, the word "do" and "anxiety" have bad separations between Low-Medium-High risk attribution distribution, as such, grouping these modalities into 1 class would better model the distribution split of the attributions for these cases.

We also noted that our TF-IDF scaled mean attribution approach delivered the most balanced results between Recall and F1-score, making it a promising direction for future development and analysis. Longitudinal classification results (w.r.t. merged classes) are also encouraging, especially with large recall for No Risk - Any Risk classification. We can use our method for preliminary screening of suicidal ideation then put those results through a more robust model.

Through this work, we underscore the utility and potential of advanced machine learning techniques in mental health diagnosis and intervention. By leveraging LLMs and novel attribution methods, we believe that it is possible to significantly enhance the early detection and prevention of suicidal tendencies from social media data. However, we also acknowledge the potential ethical implications and challenges in handling sensitive information. As such, future work must ensure careful handling and privacy of data while maximizing the benefit derived from such models. We hope our research contributes to a deeper understanding of how AI can be leveraged to improve mental health care.

\section{Future Work}
Our research opens up several avenues for exploration in the domain of using AI for detecting suicidal ideation in social media data:
\begin{itemize}
    \item State-of-the-Art Comparison: This study utilized Layer Integrated Gradients with only encoder-based LLMs. A valuable next step would be to conduct an in-depth comparison of our approach with other state-of-the-art models in the field. This would not only help in benchmarking the performance but also provide insights into potential synergies or enhancements when combined with other methodologies.
    \item Multimodal Analysis: With social media platforms increasingly supporting multimedia content, analyzing associated imagery or videos could further refine detection accuracy.
    \item Longitudinal Insights: A more granular time-series analysis of users' posts might provide insights into evolving patterns of suicidal ideation or early warning signs.
\end{itemize}

By delving into these directions, future research can further harness the potential of AI to offer timely interventions and support to individuals exhibiting signs of suicidal tendencies on social media platforms.

\bibliography{anthology,icmla2023}
\bibliographystyle{IEEEtranN}



\end{document}